# CDDO–HS: Child Drawing Development Optimization–Harmony Search Algorithm

Azad A. Ameen [1,2,*], Tarik A. Rashid [3] and Shavan Askar [1]

[1] Information Systems Engineering Department, Erbil Technical Engineering College, Erbil Polytechnic University, Erbil/44001, Iraq; azad.ameen@charmouniversity.org

[2] Department of Computer Science, College of Sciences, Charmo University, Sulaymaniyah/46023, Iraq

[3] Computer Science and Engineering Department, University of Kurdistan Hewler, Erbil/44001, Iraq; tarik.ahmed@ukh.edu.krd

* Correspondence: azad.ameen@charmouniversity.org

**Abstract:** Child drawing development optimization (CDDO) is a recent example of a metaheuristic algorithm. The motive for inventing this method is children's learning behavior and cognitive development, with the golden ratio being employed to optimize the aesthetic value of their artwork. Unfortunately, CDDO suffers from low performance in the exploration phase, and the local best solution stagnates. Harmony search (HS) is a highly competitive algorithm relative to other prevalent metaheuristic algorithms, as its exploration phase performance on unimodal benchmark functions is outstanding. Thus, to avoid these issues, we present CDDO–HS, a hybridization of both standards of CDDO and HS. The hybridized model proposed consists of two phases. Initially, the pattern size (PS) is relocated to the algorithm's core and the initial pattern size is set to 80% of the total population size. Second, the standard harmony search (HS) is added to the pattern size (PS) for the exploration phase to enhance and update the solution after each iteration. Experiments are evaluated using two distinct standard benchmark functions, known as classical test functions, including 23 common functions and 10 CEC-C06 2019 functions. Additionally, the suggested CDDO–HS is compared to CDDO, HS, and six others widely used algorithms. Using the Wilcoxon rank-sum test, the results indicate that CDDO–HS beats alternative algorithms.

## 1. Introduction

The term "optimization" refers to the method used to pick the optimal solution from a pool of possibilities. Each process may be optimized and many difficult issues in management, economics, science, and technology can be cast as optimization problems [1]. Due to the computational effort required by conventional numerical optimization methods, it is most likely not possible to execute a thorough search of the optimal solution space for challenging engineering optimization problems. In the past two decades, there has been growing interest in the development of alternatives to traditional, mathematically derived, and gradient-based optimization strategies. This interest stems from a desire to discover more efficient methods for attaining optimal results [2]. In this scenario, heuristic optimization algorithms are better approaches for resolving difficult engineering optimization problems [3]. Due to this, heuristic optimization approaches have been developed, which can be applied to problems that derivative methods are unable to solve. As a result, a multitude of heuristic and metaheuristic algorithmic techniques has been created [4].

Metaheuristics are effective optimization strategies that have gained popularity in a variety of disciplines, including engineering, finance, and applied science [5,6]. The vast majority of cutting-edge metaheuristics were developed in the years leading up to the year 2000, when these algorithms were still considered to be "classical" metaheuristic algorithms [7]. The most common classical metaheuristics include genetic algorithms (GAs), introduced by Goldberg in 1988, which are a population-based approach that mimics natural selection and genetics to search for optimal solutions [8]. Kennedy and Eberhart proposed particle swarm optimization (PSO) in 1995, which uses a swarm of moving particles in the search space to find the best solution [9]. Kirkpatrick et al. introduced simulated annealing (SA) in 1983 as a probabilistic method that employs the physical process of annealing to seek out the optimal solution [10]. Tabu search (TS), introduced by Glover in 1986, is a local search-based metaheuristic that uses a taboo list to avoid returning to recently visited solutions [11]. Ant colony optimization (ACO), proposed by Dorigo and Stützle in 1992, uses a population of artificial ants that deposit pheromones to communicate and find the optimal solution [12]. Notwithstanding the successes of classical metaheuristic algorithms, novel and innovative evolutionary techniques have been developed lately. Throughout this age, metaheuristic algorithm research has resulted in the introduction of a substantial number of new metaheuristics inspired by evolutionary or behavioral processes. This new wave of metaheuristic techniques frequently delivers the best solutions for some benchmark problem sets that have yet to be addressed [13]. These metaheuristics have been applied to various optimization problems and have demonstrated their effectiveness and efficiency in finding near-optimal solutions in a reasonable amount of time [14].

Based on the criteria, there are many classifications of inspired metaheuristic algorithms [15–18]. Therefore, according to [18], five groups can be used to broadly categorize inspired metaheuristic algorithms:

1. Evolution-based methods, such as the genetic algorithm (GA) [8], memetic algorithm (MA) [19], genetic programming (GP) [8], biogeography-based optimizer (BBO) [20], and virulence optimization algorithm (VAO) [21].
2. Swarm-based methods, such as krill herd (KH) [22], fitness-dependent optimizer (FDO) [23], ant colony optimization (ACO) [12], bacterial foraging behavior (BFO) [24], particle swarm optimization (PSO) [9], cuckoo optimization algorithm (COA) [25], dolphin optimization algorithm (DOA) [26], dragonfly algorithm (DA) [27], bat algorithm (BA) [28], group search optimizer (GSO) [29], ant nesting algorithm (ANA) [30], and donkey and smuggler optimization (DSO) [31].
3. Physics-based methods, such as black hole (BH) [32], ray optimization (RO) [33], charged system search (CSS) [34], simulated annealing (SA) [10], big-bang–big-crunch (BBCB) [35], gravitational local search (GLSA) [36], central force optimization (CFO) [37], thermal exchange optimization (TEO) [38], and the water strider algorithm (WSA) [39].
4. Human-based methods, such as the harmony search (HS) [40], group learning algorithm (GLA) [41], firework algorithm (FA) [42], league championship algorithm (LCA) [43], learner performance-based (LPB) [44], interior search algorithm (ISA) [45], mine blast algorithm (MBA) [46], soccer league competition (SLC) [47], exchange market algorithm (EMA) [48], and the social-based algorithm (SBA) [49].
5. Natural disaster algorithms, such as the earthquake algorithm (EA) [50].

Child drawing development optimization (CDDO), as a human-based metaheuristic algorithm, is one of the latest metaheuristic algorithms. CDDO depends on the behavior of children learning to draw at an early age [51]. This algorithm demonstrated superior performance in locating the optimal global solution for optimization problems tested using classical benchmark functions. Its results were compared to multiple cutting-edge algorithms such as PSO (particle swarm optimization) [9], DE (differential evolution) [52], the WOA (whale optimization algorithm) [53], GSA (gravitational search algorithm) [36], and FEP (fast evolutionary programing) [54]. The effectiveness of the CDDO was evaluated. This demonstrated that the CDDO was exceptionally resilient when acquiring a new solution. It also employs a random search mechanism to change the position to find the best solution [13]. According to [51], CDDO performs well in terms of convergence time and balancing exploration and exploitation.

Although the method performs well compared to other algorithms, several restrictions affect how well it performs in particular circumstances. When an algorithm, for instance, becomes trapped in local optima, it finds a solution that is optimal within a specific region of the search space, but not necessarily the optimal solution overall. As the encircling mechanism is used in the search space, CDDO's performance in escaping from local solutions is poor, and convergence and speed are inefficient in both. This means that CDDO has issues relating to the balance between exploration and exploitation.

In addition, not improving the best solution more effectively is a problem associated with the encircling mechanism, and the CDDO exploitation phase must be enhanced to obtain better solutions [14,51].

The following are the primary motivations for hybridizing CDDO with HS in this paper:

1. Due to CDDO employing an encircling search mechanism, thus, suffering from the inability to avoid local optima.
2. CDDO performs poorly during the exploitation phase.
3. Improving the CDDO solution after each iteration is insufficient.

The authors came up with the hybridized algorithm as a solution to the CDDO concerns that were described previously, as well as the standard harmonic search algorithm's great exploitability in the execution of multimodal benchmark functions. As a consequence of this, the authors decided to go with a hybrid approach that combines CDDO and global HS to provide higher performance in the exploitation phase with global HS, particularly when tested using unimodal benchmark functions. Thus, the purpose of this paper is to propose a hybridized approach to overcoming CDDO problems by employing two effective mechanisms: The first mechanism for improving CDDO performance is updating the pattern memory iteratively by using the standard HS mechanism during the exploitation phase and comparing new experiences (pattern memory) with the current child's drawings. The second is to save the best solution for each iteration and then compare each new solution to the best solution in the exploration phase. If the outcome is better than the best solution, the child's drawings are changed; otherwise, they remain in their current drawings. This hybridization combines two completely different algorithm mechanisms; CDDO and HS, and adds an update condition during the exploitation phase. The child's drawings can then be updated during the exploitation phase, utilizing the HS approach. As a result, CDDO–HS is a newly proposed hybridization that enhances CDDO's performance.

This paper is organized as follows: In Sections 2 and 3, we sequentially describe CDDO and HS with their mechanisms. Section 4 describes the novelty and contribution of our work. Section 5 presents our proposed approach, CDDO–HS, including a detailed description of the hybridized algorithm. We provide experimental results and analyses in Section 6, assessed with two benchmark test functions (the classical and CEC 2019 benchmark functions), and compare it to other recent algorithms, such as FOX (FOX-inspired optimization algorithm) [55], Choa (chimp optimization algorithm) [56], BOA (butterfly optimization algorithm) [57], DCSO (dynamic cat swarm optimization algorithm) [58], WOA–BAT [59] (hybrid WOA (whale optimization algorithm) [53] with BAT (bat algorithm optimization) [28]), and GWO–WOA [60] (hybrid GWO (grey wolf optimization) [61] with WOA (whale optimization algorithm) [53]). Finally, Section 7 provides the conclusions and prospective research recommendations.

## 2. CDDO

CDDO was developed by Sabat Abdulhameed and Tarik A. Rashid in 2021 as a metaheuristic strategy for solving optimization issues with a single object. CDDO takes inspiration from children's natural tendencies to learn and grow intellectually and applies the principles of the golden ratio to their work to bring out their maximum capabilities for beauty [14]. CDDO uses the golden ratio and mimics cognitive development and the steps a youngster takes to improve from unskilled scribbling to proficient pattern drawing. The "golden ratio", a mathematical relationship between any two consecutive numbers in the Fibonacci sequence, is ubiquitous in nature, art, architecture, and design. If a child's hand pressure is adjusted for width, length, and the golden ratio, the resulting drawing becomes more aesthetically beautiful. This fosters a child's natural development, raises their intelligence, and teaches them how to work together towards a common goal [14,51]. Table 1 shows the formulation of CDDO optimization [51].

*Table 2 The formulation of CDDO optimization*

| Statement | Description |
|---|---|
| f(x) | The cost function to obtain the best drawing. |
| Xij | The current solution is *X*, which is a child's drawing that is influenced by factors such as hand pressure, the golden ratio, length, and width. *i* represents the number of decision variables (the population number). *j* represents the number of dimensions that differ. |
| PM | Pattern memories are abilities that are acquired through experience. They use the feedback to identify patterns in the drawings, strive to give the pictures meaning, and exercise through copying, practicing, and being enthusiastic (with a trial). |
| GR | The ratio between a child's drawing's length and width, serving as the two components of the solution, is known as the golden ratio. |
| HP | Current solutions exert pressure between the lower boundary (LB) of the problem and the upper boundary (UB) of the solution (UP). |
| RHP | The children's drawings each have hand pressure, and one of those hand pressures is chosen at random. |
| LB, UB | The lower boundary (LB) of the problem and the upper boundary (UB) of the problem are the width and length of a child's artwork, respectively. |
| RAND | The expression RAND is used to symbolize the action of producing a random number from a set of parameters. |

This algorithm is divided into five phases [51]:

- Stage one: The scribble initialization, as shown below.

$$X_{(i,j)} = \begin{bmatrix} X_{(1,1)} & X_{(1,2)} & \cdots & X_{(1,j)} \\ X_{(2,1)} & X_{(2,2)} & \cdots & X_{(2,j)} \\ X_{(3,1)} & X_{(3,2)} & \cdots & X_{(3,j)} \\ \vdots & \vdots & \cdots & \vdots \\ X_{(i,1)} & X_{(i,1)} & & X_{(i,j)} \end{bmatrix} \quad (1)$$

In the equation above, *i* stands for the population size and *j* stands for the number of variable dimensions. Within the allowed range, all population member dimensions can be set freely.

- Stage two: Exploitation

This stage teaches the child to control the movement and direction. Equation (2) generates the RHP. The RHP is a random number that is used to evaluate the current solution's hand pressure (HP). This number is located between the lower boundary of the problem (LB) and the upper boundary of the solution (UP). Where HP denotes the pressure applied with the hand and j refers to the parameters of the solution, Equation (3) can select the HP.

$$\text{RHP} = \text{RAND}(\text{LB(Lower Boundary)}, \text{UP (Upper Boundary)}) \quad (2)$$

$$\text{HP} = X_{(i,\text{RAND}(j))} \quad (3)$$

(i,RAND(j)) represents the current solution's **i** hand pressure among several drawing solutions **(j)**. RAND generates a random number between two or more variables.

- Stage three: The golden ratio

The golden ratio (GR) is also utilized to update and enhance the efficacy of the solution. The ratio of a child's artwork's width to its length is one of the elements that are considered when determining how to solve a problem (see Equation (4)). Using Equation (5), each of these two elements can be chosen at random from among all the problem factors.

$$XGR_i = \frac{XL_{(i,M)} + XW_{(i,N)}}{XL_{(i,M)}} \tag{4}$$

To calculate the golden ratio, which is based on a random selection of two drawings for each population **(i),** we first added the first and second drawings together and then divided the result by the second drawing using Equation (4).

$$XL_{(i,M)}, XW_{(i,N)} = X_{(i,RAND(j))}, \ XL_{(i,M)} \neq XW_{(i,N)} \tag{5}$$

To ensure that the first and second drawings were different, we utilized Equation (5) to choose the drawings (dimension) (j) randomly from a population (i).

The child would now use the knowledge and abilities they gained from previous experiences and criticisms by trying to analyze the patterns in the actual pictures, attempting to give the drawings some sort of significance, and honing their drawing skills through copying, practicing, and being passionate about their work (with trial).

To start implementing these behaviors, first, the child's competency can be measured by having them measure their hand pressure (HP). If the current hand pressure is lower than the RHP, the outcome can be recalculated using Equation (3), which takes into consideration both the child's skill rate (SR) and level rate (LR).

In Equation (6), $X_i lbest$ is a child's best drawing thus far, and $X_i gbest$ is the children's consensus on the globe's best solution, as determined by the conditions in their surroundings. In addition, as mentioned previously, the golden ratio (GR) is the proportion of the length (L) to the width (W) of a child's drawing (W).

$$X_{i+1} = GR + SR * (X_i lbest - X_i) + LR * (X_i gbest - X_i) \tag{6}$$

- Stage four: Creativity

Children gain creativity and skills through experience and observation. Every artwork benefit from creativity. At this level, the child revises the golden ratio solutions. Regrettably, there is no meaningful hand pressure in the answer, indicating that a child's talents are not yet established and need to be improved upon by utilizing the creative factor and golden ratio. Every child remembers the best learning methods and attempts to reproduce them to progress. Each algorithm solution has a pattern memory (PM), the size of which is determined by the problem. Using a random solution from the PM array to update underperforming solutions can boost the algorithm's convergence rate and accelerate children's learning for a long period.

Equation (7) uses CR and PM to update the current solution and converge to the ideal solution. The creative factor, CR = 0.1, improves performance. Later on, the child's SR and LR can both be set to a low value (between 0 and 0.5), indicating that the kid has a poor skill rate and inaccurate knowledge, but that their originality and pattern memory can grow.

$$X_{i+1} = X_i PM + SR * (X_i gbest) \tag{7}$$

- Stage five: Pattern Memory

This pertains to adding more specifics and increasing the level of precision, as well as comparing the results to the most accurate drawings possible while drawing on prior experience and expertise. The current drawing can have an accurate golden ratio, but be unaffected by the hand pressure of the user because the algorithm randomly selects one of the child's best drawings to use as the basis for the update. This stage emphasizes drawing in finer details. The algorithm applies the agent's own optimal updating mechanism's behavior. If better solutions exist, the population's global best solution would be revised. This also applies to updating the pattern memory with each iteration's best global answer. CDDO's pseudocode demonstrates how the CDDO algorithm works [51] (see Algorithm 1):

**Algorithm 1:** Child drawing development optimization (CDDO) algorithm

```
Start
    Create a population of children's drawings Xi (i = 1, 2, ..., j)
    Calculate the fitness of each drawing, then create personal and global benchmarks.
    Compute each drawing's golden ratio using Equation (4)
    Create a pattern memory (PM) array
    Select a pattern memory index at random
    If (t < maximum number of iterations)
            Utilize Equation (2) to calculate RHP
            Choose at random the hand pressure using Equation (3)
                    If (hand pressure (HP) is low)
                            Utilize Equation (6) to update the drawings
                            Set LR and SR to HIGH (0.6–1)
                    Else if
                            XiGR is near to golden ratio (GR)
                            Consider the learned patterns, LR and SR using
                            Equation (7)
                            Set LR and SR to low (0–0.5)
                    End if
                    Evaluate the fitness function of cost values
                    Update local (personal) best
                    Update global best
                    Update pattern memory (PM)
                    Store the best cost value
        Increase t
    End if
    Return global best
End
```

## 3. HS

The harmony search algorithm is a metaheuristic optimization technique inspired by the process of musical improvisation, where a musician generates a melody that follows a certain harmony. The HS algorithm was first proposed by Geem et al. in 2001 [40] and has since been applied to solve various optimization problems in engineering, finance, and other fields. The HS algorithm simulates a group of musicians who generate new melodies by improvising and adjusting their playing style based on their previous experiences and the harmony of the melody. The algorithm searches for the optimal solution by adjusting the harmony among the variables of the problem. The HS algorithm is effective in solving complex optimization problems, especially in situations where the problem space is continuous and high-dimensional. The HS algorithm has been used in various applications, such as design optimization, image processing, and signal processing. Several modifications and extensions of the HS algorithm have also been proposed to improve its performance and applicability [62–65].

The initialized population members of the HS algorithm are independent in each dimension and fall within the allowable range. The algorithm only creates one new member in each iteration. Then, using either a memory consideration rule and a pitch adjustment factor or all random reinitializations in the permitted range of dimensions, each dimension of the new point is generated from all of the solutions in the HM. The population member with the highest cost function value is compared to the new generation of solutions, and if the new solution has a lower cost, the population member is replaced. Up until one of the termination criteria is satisfied, this process is repeated [40,66].

The following is a description of the HS algorithm sequence [67,68]:

1. Define the cost function (**f (x)**) that needs to be reduced to achieve the algorithm's objective.

2. Set the parameters as shown below to begin.

$$HM = \begin{bmatrix} X_1^1 & X_2^1 & \cdots & X_j^1 \\ X_1^2 & X_2^2 & \cdots & X_j^2 \\ X_1^3 & X_2^3 & \cdots & X_j^3 \\ \vdots & \vdots & \cdots & \vdots \\ X_1^{HMS} & X_2^{HMS} & & X_j^{HMS} \end{bmatrix} \quad (8)$$

The population number in the above equation is the HS memory (HMS) and the number of variable dimensions is (j). All population member dimensions can be set at random within the authorized range. The initial values of the HS consideration rate (HMCR) and pitch adjustment rate (PAR) are often set to (0.995) and (0.1).

3. Generate a new point ($X^{new} = X_1^{new}, X_2^{new}, X_3^{new}, \ldots, X_j^{new}$) by performing the following:

A corresponding member dimension is chosen at random using HMCR for each of the n dimensions. The value of the new point is chosen at random from the authorized range:

$$X_i^{new} = \begin{cases} X^{new}(i) \in \{X_j^1, X_j^2, X_j^3, \ldots, X_j^{HMS}\} \; if \; rand(0,1) \leq HMCR \\ X^{new}(i) \; is \; random \; if \; it \; isn't \end{cases} \quad (9)$$

$$X_i^{new} = X^{new}(i) + \text{RAND}(-1,1) \times bw, if \; rand(0,1) \leq PAR$$
$$\text{where bw} = 0.04 \quad (10)$$

4. If the new harmony vector $X^{new}$ has a lower cost, replace the worst member of the population with it.
5. Verify the termination criteria; if they are satisfied, move on to step three; otherwise, the optimum point is identified. The HS pseudocode demonstrates how the HS algorithm works [69] (see Algorithm 2):

---

**Algorithm 2:** Harmony search (HS) algorithm
**Begin**
    Define objective function f(x), x = (x, x2, …, xa)
    Define harmony memory considering rate (HMCR)
    Define pitch-adjusting rate (PAR) and other parameters
    Generate harmony memory with random harmonies
    **While** (t<max number of iterations)
        **While** (≤number of variables)
            **If** (RAND<HMCR)
                Choose a value from HM for variable **i** (Equation (9))
                **If** (RAND<PAR)
                      Adjust the value by adding a certain amount (Equation (10))
                **End if**
            **Else**
                Choose a random value
            **End if**
        **End while**
        Accept the new harmony (solution) if better
    **End while**
    Find the current best solution
**End**

---

## 4. The Novelty and Contribution

The goal of our study is to introduce a novel hybrid metaheuristic algorithm for optimization, which combines two existing approaches: child drawing development optimization (CDDO) and global harmony search (HS). By combining the best features of CDDO and HS, the suggested hybrid algorithm, CDDO–HS, can improve upon the optimization process. Our most significant achievement thus far has been the creation of a novel hybridization method for the efficient joint use of CDDO and conventional HS. To be more specific, we suggest a new hybridization strategy that integrates CDDO's local search capabilities with traditional HS's worldwide exploration potential. The solution space can be efficiently searched with this method, resulting in faster convergence and better solutions overall.

The suggested CDDO–HS algorithm is also subjected to a thorough performance evaluation on a variety of benchmark optimization tasks. In terms of solution quality, convergence speed, and robustness, the CDDO–HS algorithm is shown to excel in experimental settings, compared to both CDDO and baseline HS. The CDDO–HS algorithm outperforms many other state-of-the-art metaheuristic algorithms, including some that were developed relatively recently. These include the FOX (FOX-inspired optimization algorithm), Choa (chimp optimization algorithm), BOA (butterfly optimization algorithm), DCSO (dynamic cat swarm optimization algorithm), and GWO–WOA (hybrid grey wolf optimization with whale optimization algorithm).

**5. CDDO–HS**

Based on the previous sections on CDDO and the standard HS, the proposed approach was described in this section by combining CDDO and HS to improve the performance of CDDO in terms of efficiency during the exploitation phase to produce gain advantages. In general, the standard CDDO is capable of locating the optimal solution. Unfortunately, it is not sufficient in refining the optimal solution with each iteration. Hence, CDDO was hybridized with global HS to enhance its performance. The name of the hybridization algorithm was decided as CDDO–HS. Thus, the CDDO was hybridized by including two strategies:

First, we moved the pattern size (PS) to the algorithm's core, where it changed with each iteration. The CDDO algorithm assumed that drawing skills from childhood to adolescence are stored in memory once and do not need to be refreshed, which is incorrect. As a result, children's drawings are regenerated as they draw and throughout their life, which means that the pattern size is constantly renewed and aids the child in the subsequent stages of drawing. In each sketching session, the child learns new skills that are far superior to and distinct from previous sessions.

To achieve the best exploration, we set the pattern size to 80% of the overall population size when we started. On the other hand, we attempted to update the pattern size of the algorithm that would power CDDO using a mechanism or technique. We used the global harmony search technique to update the pattern size after multiple experiments and different algorithms. Moreover, the harmony search algorithm offered various advantages, such as its balance of exploration and exploitation, adaptability, simplicity, and resilience. The HS hybridization with other algorithms could assist in overcoming its limits and improving its performance.

These modifications resulted in improved performance for obtaining the optimum fitness function. Algorithm 3 and Figure 1 show the CDDO–HS pseudocode and flowchart, respectively.

---

**Algorithm 3** Child-drawing-development-optimization–harmony-search-based hybrid algorithm

**Start**

    Create a population of children's drawings Xi (i = 1, 2, ..., j)

    Calculate the fitness of each drawing, then create personal and global benchmarks.

    Compute each drawing's golden ratio using Equation (4)

Create an array of pattern memory (PM)
            Select a pattern memory index at random
            **If** (t < maximum number of iterations)
                    **If** (i ≤ number of variables)
                            **If** (RAND<HMCR)
                                    Choose a value from updated pattern memory for the variable, using Equation (9)
                                            **If** (RAND<PAR)
                                                Increase the value by a given amount, using Equation (10)
                                            **End if**
                            **Else**
                                Choose a random value
                            **End if**
                    **End if**
                    Update the pattern memory
                    Utilize Equation (2) to calculate the RHP
                    Choose at random the hand pressure using Equation (3)
                                    **If** (hand pressure (HP) is low)
                                            Utilize Equation (6) to update the drawings
                                            Set LR and SR to high (0.6–1)
                                    **Else if**
                                            XiGR is near the golden ratio (GR)
                                            Consider the learned patterns, LR and SR using Equation (7)
                                            Set LR and SR to low (0–0.5)
                                    **End if**
                                    Evaluate the fitness function of cost values
                                    Update local (personal) best
                                    Update global best
                                    Update pattern memory (PM)
                                    Store the best cost value
                    Increase t
            **End if**
            Return global best
**End**

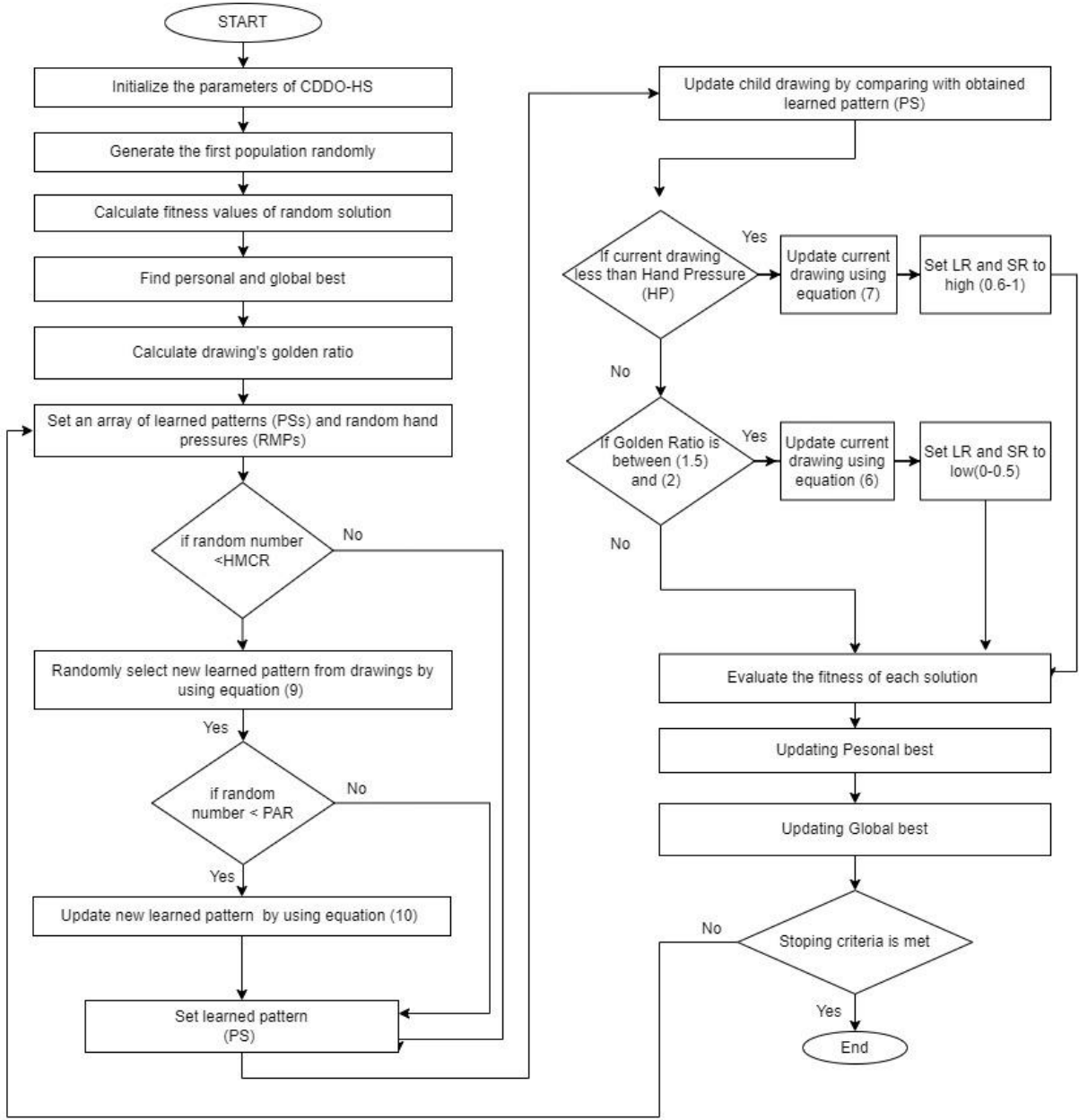

**Figure 1.** Child-drawing-development-optimization–harmony-search-based hybrid flowchart.

## 6. Testing, Results, and Discussion

To verify the efficacy of a novel method and determine meaningful comparisons to state-of-the-art and classical algorithms, optimization researchers employ a conventional approach. In optimization research, this method is frequently employed to assess candidates for optimization on a sizable test set. Therefore, the CDDO–HS method was implemented and examined with 23 classical benchmark functions and 10 benchmark functions from CEC 2019. The parts that follow detail the CDDO–HS's evaluations in comparison to other metaheuristic algorithms, as well as its benchmark functions, experimental results, evaluations, and statistical data.

*6.1. Benchmark Functions*

To ensure the accuracy of our proposed CDDO–HS, we tested it using two sets of standard benchmark functions. As an initial standard, we used a set of 23 different classical functions. The CEC-C06 2019 standard was the benchmark function's second component. Some examples of these operations include the multimodal function, the unimodal function, the expanding multimodal function, and the hybrid composition operation. You can view these standard test functions in [23].

*6.2. Experimental Setup*

MATLAB R2020a was used to implement the code on Windows 11. To achieve a better and more accurate outcome, the first population was randomly selected. The parameter initialization for the implementation was set as follows:

1. Population size = 40.
2. The number of iterations = 500.
3. Run of Algorithms= 30.

*6.3. Evaluation Criteria*

CDDO–HS could be evaluated in a variety of ways. The following were the evaluation points:

1. The average and standard deviations were displayed.
2. When comparing CDDO–HS to CDDO.
3. When comparing CDDO–HS to HS.
4. CDDO–HS vs. other metaheuristic algorithms (FOX, Choa, DCSO, WOA–BAT. and WOAGWO).
5. Making a box and whisker plot to compare CDDO, HS, and CDDO–HS.

*6.4. CDDO–HS vs. CDDO and HS*

To validate the performance of CDDO–HS, two types of benchmark functions were used: classical benchmark functions and CEC-C06 2019. Hence, the obtained results' average (Ave), probability value ($p$-value), and standard deviation (Std) were utilized to evaluate the performance testing.

6.4.1. Classical Benchmark Functions

The classical benchmark functions are a set of standard functions used to evaluate the performance of optimization algorithms. These functions are commonly used in the field of optimization and have been studied extensively [70]. Functions f1–f7 are called unimodal functions because they only have one global optimum and no local optimums. They have a single solution and can be exploited as variants (see Table A). Therefore, using these unimodal functions, it was possible to assess the CDDO–HS exploitation capabilities. Table 2 demonstrates that compared to the CDDO in these functions, CDDO–HS had improved exploitation capabilities in two of the seven. We also discovered that all seven equations had significant outcomes when we compared the results of this hybrid algorithm to those of the HS algorithm. The results of the CDDO algorithm's transitions were denoted with boldface type, while those of the HS algorithm's transitions were denoted with underlining.

**Table 2.** Table comparing CDDO–HS with CDDO and standard HS.

| Fun | CDDO–HS | | CDDO | | HS | |
|---|---|---|---|---|---|---|
| | Avg. | Std. | Avg. | Std. | Avg. | Std. |
| F1 | 5.087E-33 | 1.057E-32 | **1.328E-57** | **8.635E-73** | 2.850E+02 | 9.021E+01 |
| F2 | 4.921E-17 | 4.033E-17 | **2.453E-32** | **4.385E-32** | 3.005E+00 | 5.355E-01 |
| F3 | 1.249E-29 | 2.238E-29 | **2.736E-39** | **5.168E-40** | 1.754E+04 | 5.815E+03 |
| F4 | 1.986E-16 | 1.920E-16 | **7.815E-33** | **2.784E-48** | 2.214E+01 | 1.787E+00 |
| F5 | **2.298E+00** | **6.935E+00** | 2.419E+01 | 1.023E+01 | 2.040E+04 | 8.759E+03 |
| F6 | **5.589E-04** | **1.439E-04** | 7.074E-01 | 6.787E-01 | 2.834E+02 | 1.023E+02 |
| F7 | 2.901E-03 | 1.497E-03 | **1.361E-03** | **1.123E-03** | 2.042E-01 | 5.145E-02 |
| F8 | -1.178E+04 | 3.098E+03 | **-1.244E+04** | **5.537E+02** | -1.240E+04 | 7.638E+01 |
| F9 | **2.222E+00** | **6.070E+00** | 1.060E+01 | 1.724E+01 | 1.968E+01 | 3.683E+00 |
| F10 | **6.809E-15** | **1.703E-15** | 7.875E-15 | 4.118E-15 | 5.095E+00 | 5.702E-01 |
| F11 | **0.000E+00** | **0.000E+00** | 5.688E-01 | 1.532E+00 | 3.513E+00 | 8.925E-01 |
| F12 | **1.161E-06** | **3.112E-07** | 3.167E-01 | 9.0046E-01 | 7.060E+00 | 2.147E+00 |
| F13 | **<u>1.555E-05</u>** | **<u>5.271E-06</u>** | 4.128E-01 | 3.745E-01 | 1.622E+02 | 1.596E+02 |
| F14 | 5.372E+00 | 4.241E+00 | **9.981E-01** | **3.686E-04** | <u>9.980E-01</u> | <u>3.448E-11</u> |
| F15 | **<u>6.249E-04</u>** | **<u>2.502E-04</u>** | 1.181E-03 | 1.022E-03 | 6.699E-03 | 9.110E-03 |
| F16 | **<u>-1.032E+00</u>** | **<u>3.777E-10</u>** | -1.029E+00 | 3.247E-03 | -1.032E+00 | 1.889E-07 |
| F17 | **<u>3.979E-01</u>** | **<u>2.675E-09</u>** | 4.231E-01 | 4.814E-02 | 3.979E-01 | 6.467E-06 |
| F18 | 5.700E+00 | 8.238E+00 | **3.117E+00** | **1.579E-01** | <u>3.900E+00</u> | <u>4.930E+00</u> |
| F19 | **<u>-3.863E+00</u>** | **<u>1.554E-10</u>** | -3.728E+00 | 1.075E-01 | -3.863E+00 | 4.714E-08 |

To evaluate our suggested algorithm's performance in terms of exploration, we used multimodal functions ranging from F8 to F13 (see Tables 1 and A2) to analyze the local optimum avoidance. The ability to start from the local optima and continue the search across a wide range of regions of the search area could be tested by utilizing multimodal features in an exploration capacity test. To put it another way, they ranked the depth of the variation investigation for a broad set of local optimums. Five out of six multimodal functions (Table 2) suggested that CDDO–HS performed better. Hence, it could be asserted that CDDO–HS enhanced CDDO's exploratory capabilities. Consequently, when the results of this method were compared to the results of the HS algorithm, six out of seven outcomes were significant.

The third section of these benchmark functions included fixed-dimension functions ranging from F14 to F19 (see Tables 4 and A3). Hence, avoiding the local minima was one of the most difficult challenges, because only by finding an optimal balance between exploration and exploitation could a local minimum be avoided to some extent. Due to a large number of local minima, the multimodal fixed-dimension benchmark functions were employed to test the algorithm's ability to avoid them. The numerical results of the benchmark functions from F14 to F19 showed that CDDO–HS outperformed the CDDO and standard HS algorithms in terms of local minima avoidance, except for F14 and F18, where the CDDO–HS algorithm performed worse than the other algorithms. After a while, CDDO–HS was shown to compete with other cutting-edge metaheuristic algorithms, as evidenced by its remarkable findings.

This showed that the algorithm could avoid the local minima because it looked at a lot of good places in the design space and chose the best one. The method used to avoid the local minima was that all of the search agents would change suddenly in the early stages of the optimization process and then progressively converge toward the best solution. This strategy later ensured that cooperative search agents eventually converged on a spot in the search space.

6.4.2. CEC-C06 2019 Benchmark Test Functions

CEC-C06 2019 Benchmark Test Functions are a set of 10 mathematical functions commonly used to evaluate the performance of optimization algorithms. These functions were specifically designed for the IEEE Congress on Evolutionary Computation (CEC) 2019 competition to provide a more challenging and diverse set of test functions compared to previous years. These functions cover a wide range of optimization problems, including unimodal, multimodal, separable, and non-separable functions. Some of the functions also incorporate features such as rotation, scaling, and shifting to further increase their complexity. Each function is defined over a set of input variables and has a known global minimum or maximum, which allows for optimization algorithms to be evaluated based on their ability to find the optimal solution. Thus, they are widely used in the field of optimization and evolutionary computation to compare the performance of different algorithms and to benchmark new algorithms. They are also used to evaluate the effectiveness of various optimization techniques, such as metaheuristics, swarm intelligence, and genetic algorithms [71].

The CDDO–HS algorithm was also evaluated using the CEC-C06 2019 benchmark function. Table 3 and Figure 2 demonstrate that CDDO–HS outperformed CDDO in all functions. Similarly, when we compare the outcomes of this approach to the standard HS algorithm, we found that CDDO–HS outperformed the HS algorithm in all except three functions: F4, F5, and F6. Overall, CDDO–HS outperformed CDDO and standard HS in eight multimodal benchmark functions.

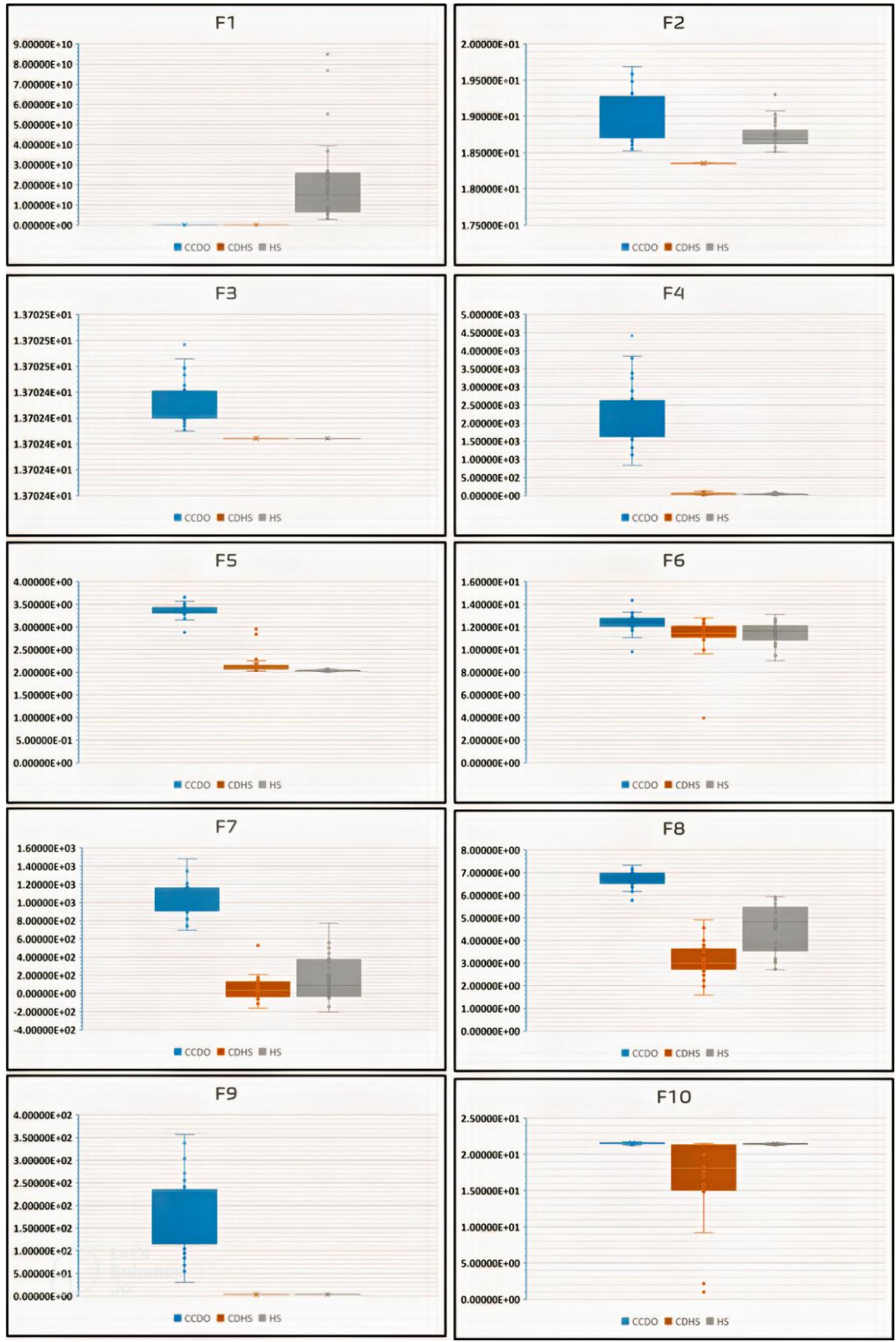

**Figure 2.** The CDDO, HS, and CDDO–HS box and whisker plots for CEC-C06 2019.

**Table 3.** Table comparing CDDO–HS with CDDO and standard HS on CEC-C06 2019.

| Fun | CDDO–HS | | CDDO | | HS | |
|---|---|---|---|---|---|---|
| | Avg. | Std. | Avg. | Std. | Avg. | Std. |
| F1  | **5.317E+04** | **8.142E+03** | 4.104E+05 | 3.377E+05 | 2.232E+10 | 2.295E+10 |
| F2  | **1.835E+01** | **6.390E-03** | 1.898E+01 | 3.663E-01 | 1.875E+01 | 2.090E-01 |
| F3  | **1.370E+01** | **5.785E-13** | 1.370E+01 | 1.633E-05 | 1.370E+01 | 4.987E-10 |
| F4  | **5.746E+01** | **3.066E+01** | 2.221E+03 | 8.515E+02 | 3.976E+01 | 2.003E+01 |
| F5  | **2.170E+00** | 2.109E-01 | 3.350E+00 | **1.421E-01** | 2.039E+00 | 2.623E-02 |
| F6  | **1.130E+01** | 1.591E+00 | 1.238E+01 | **7.803E-01** | 1.139E+01 | 1.091E+00 |
| F7  | **5.440E+01** | **1.300E+02** | 1.040E+03 | 1.841E+02 | 1.708E+02 | 2.455E+02 |
| F8  | **3.150E+00** | **7.640E-01** | 6.710E+00 | **3.515E-01** | 4.543E+00 | 1.082E+00 |
| F9  | **3.484E+00** | **1.912E-01** | 1.790E+02 | 8.510E+01 | 3.634E+00 | 1.742E-01 |
| F10 | **1.554E+01** | 7.609E+00 | 2.155E+01 | **1.127E-01** | 2.146E+01 | 1.050E-01 |

Figures 3 and Figure 4 show the simulation results comparing CDDO–HS to CDDO and the standard HS method. The simulated convergence curves for the accepted testing functions showed that, when compared to CDDO and the standard HS algorithm, the convergence velocity and optimization precision of CDDO–HS was the best.

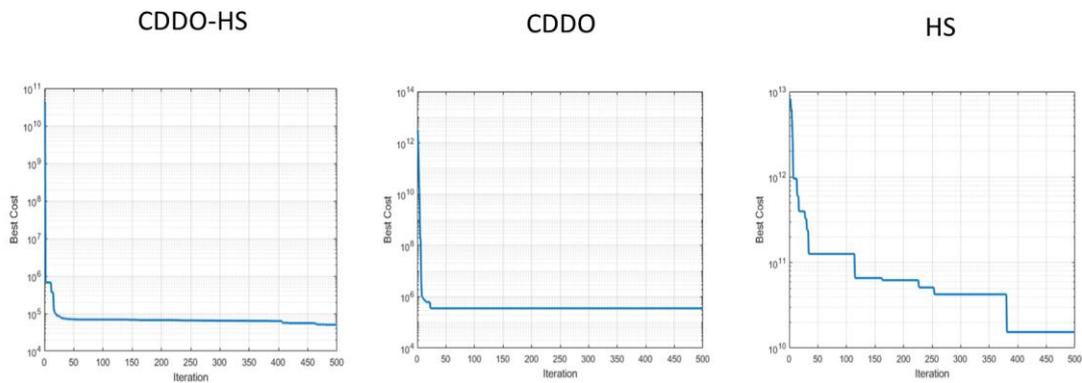

**Figure 2:** Convergence curves for function F1.

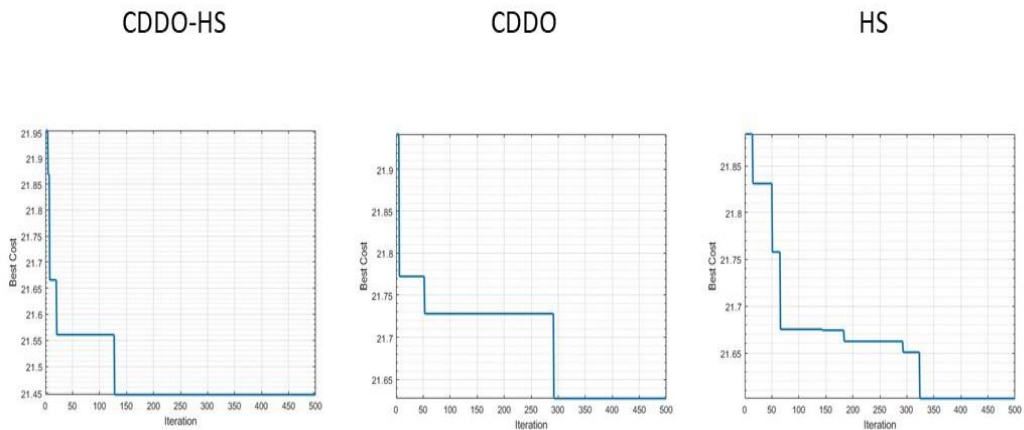

**Figure 3:** Convergence curves for function F10.

*6.5. Statistical Analysis*

To figure out if the performance of the algorithm was significant or not, we needed to use statistical analysis. Therefore, we used the p-value to figure out how important a result was from a statistical point of view and to decide whether or not to reject the null hypothesis.

To show whether the results were significant or not, Table 4 was used to find the p-values for both benchmark tests: the classical benchmark and the CEC-C06 2019 benchmark functions.

**Table 4.** The P-value of CDDO–HS against CDDO for classical benchmark functions and CEC-C06 2019.

| Function | Classical Benchmark Functions | CEC-C06 2019 |
|---|---|---|
| 1 | - | **2.99373E-07** |
| 2 | - | **3.10921E-13** |
| 3 | - | **3.94470E-13** |
| 4 | - | **3.97091E-20** |
| 5 | **9.44415E-14** | **3.87125E-33** |
| 6 | **4.16584E-07** | **1.43887E-03** |
| 7 | - | **9.09047E-32** |
| 8 | 2.54526E-01 | **5.07244E-31** |
| 9 | **1.48536E-02** | **2.82731E-16** |
| 10 | **5.58851E-02** | **6.02103E-05** |
| 11 | **4.66181E-02** | |
| 12 | **5.88921E-02** | |
| 13 | **1.18387E-07** | |
| 14 | - | |
| 15 | **5.29812E-03** | |
| 16 | **1.31944E-05** | |
| 17 | **5.69975E-03** | |
| 18 | 9.13712E-02 | |
| 19 | **4.87510E-09** | |

The results showed that CDDO–HS performed better than CDDO in 11 of the classical benchmark's nineteen functions. Furthermore, the findings showed that CDDO–HS outscored CDDO in all 10 functions of the CEC 2019 benchmark.

Table 5 was utilized to generate the p-values for both benchmark tests, the classical benchmark and the CEC-C06 2019 benchmark functions, while we compared CDDO–HS to standard HS to demonstrate if the results were significant or not.

**Table 5.** The P-value of CDDO–HS against HS for classical benchmark functions and CEC-C06 2019.

| Function | Classical Benchmark Functions | CEC-C06 2019 |
|---|---|---|
| 1 | **1.41157E-24** | **1.69507E-06** |
| 2 | **1.38125E-37** | **6.92112E-15** |
| 3 | **1.31874E-23** | **5.89563E-08** |
| 4 | **5.90033E-57** | - |
| 5 | **1.80505E-18** | - |
| 6 | **7.63997E-22** | 7.89851E-01 |
| 7 | **3.13914E-29** | **2.54399E-02** |
| 8 | 2.75280E-01 | **3.36825E-07** |
| 9 | **1.68215E-19** | **2.36583E-03** |
| 10 | **7.24212E-49** | **7.52043E-05** |
| 11 | **2.24778E-29** | |
| 12 | **1.98825E-25** | |
| 13 | **6.98511E-07** | |
| 14 | - | |
| 15 | **5.62412E-04** | |
| 16 | **2.21129E-05** | |
| 17 | **3.26857E-02** | |

|    |              |
| -- | ------------ |
| 18 | 3.08727E−01  |
| 19 | **3.55345E−04** |

Furthermore, the results showed that CDDO–HS beat standard HS in 16 functions, as opposed to 19 in the classic benchmark functions. As a result, the findings showed that CDDO–HS outperformed regular HS in seven of the CEC-C06 2019 benchmark functions.

*6.6. The CDDO-CD Algorithm in Context with Hybrid and Metaheuristic Approaches*

Using the CEC2019 test functions, the CDDO–HS hybrid algorithm was contrasted with six new hybrid and metaheuristic approach algorithms (ChOA, BOA, FOX, GWO–WOA, WOA–BAT, and DCSO).

According to Table 6, the gardened algorithm outperformed all six metaheuristic approach algorithms in six out of ten benchmarks (CEC04, CEC05, CEC07, CEC08, CEC09, and CEC10), while five algorithms yielded 1.370E+01 in CEC03.

**Table 6.** Table comparing CDDO–HS with ChOA, BOA, FOX, GWO–WOA, WOA–BAT, and DCSO on CEC-C06 2019.

| Function | CDDO–HS | ChOA | BOA | FOX | WOAGWO | WOA–BAT | DCSO |
| --- | --- | --- | --- | --- | --- | --- | --- |
| F1 | 5.317E+04 | 4.240E+09 | 5.890E+04 | **2.580E+04** | 4.760E+04 | 7.600E+07 | 3.863E+04 |
| F2 | 1.835E+01 | 1.841E+01 | 1.890E+01 | 1.834E+01 | 1.834E+01 | **1.750E+01** | 1.834E+01 |
| F3 | 1.370E+01 | 1.370E+01 | 1.370E+01 | 1.370E+01 | 1.370E+01 | **1.270E+01** | 1.370E+01 |
| F4 | **5.746E+01** | 5.933E+03 | 2.090E+04 | 1.060E+03 | 2.537E+02 | 2.120E+03 | 7.266E+01 |
| F5 | **2.170E+00** | 4.209E+00 | 6.180E+00 | 5.315E+00 | 2.426E+00 | 2.440E+00 | 2.493E+00 |
| F6 | 1.130E+01 | 1.215E+01 | 1.180E+01 | **5.033E+00** | 1.137E+01 | 1.110E+01 | 8.864E+00 |
| F7 | **5.440E+01** | 1.007E+03 | 1.040E+03 | 3.068E+02 | 5.876E+02 | 6.060E+02 | 3.291E+02 |
| F8 | **3.150E+00** | 6.785E+00 | 6.340E+00 | 5.462E+00 | 5.587E+00 | 5.720E+00 | 5.160E+00 |
| F9 | **3.484E+00** | 4.493E+02 | 2.270E+03 | 3.796E+00 | 5.671E+00 | 2.280E+01 | 6.104E+00 |
| F10 | **1.554E+01** | 2.150E+01 | 2.150E+01 | 2.098E+01 | 2.156E+01 | 2.120E+01 | 2.113E+01 |

However, when contrasting each of the six algorithms with CDDO–HS separately, CDDO–HS gave superior results, and we used the ranking sort to encapsulate the findings (see Table 7).

**Table 7.** The ranking sort of CDDO–HS, ChOA, BOA, FOX, GWO–WOA, WOA–BAT, and DCSO algorithms.

| Function | 1st | 2nd | 3rd | 4th | 5th | 6th | 7th |
| --- | --- | --- | --- | --- | --- | --- | --- |
| F1 | FOX | DCSO | GWO–WOA | CDDO–HS | BOA | WOA–BAT | ChOA |
| F2 | WOA–BAT | GWO–WOA | FOX | DCSO | CDDO–HS | ChOA | BOA |
| F3 | WOA–BAT | BOA | ChOA | DCSO | DCSO | CDDO–HS | FOX |
| F4 | CDDO–HS | DCSO | GWO–WOA | FOX | WOA–BAT | ChOA | BOA |
| F5 | CDDO–HS | GWO–WOA | WOA–BAT | DCSO | ChOA | FOX | BOA |
| F6 | FOX | DCSO | WOA–BAT | CDDO–HS | GWO–WOA | BOA | ChOA |
| F7 | CDDO–HS | FOX | DCSO | GWO–WOA | WOA–BAT | ChOA | BOA |
| F8 | CDDO–HS | DCSO | FOX | GWO–WOA | WOA–BAT | BOA | ChOA |
| F9 | CDDO–HS | FOX | GWO–WOA | DCSO | WOA–BAT | ChOA | BOA |
| F10 | CDDO–HS | FOX | DCSO | WOA–BAT | ChOA | BOA | GWO–WOA |

The ranking score could be used to evaluate the performance of various algorithms and determine which one would be most effective for a particular task. In this case, better performance was indicated by a lower-ranking score, whereas worse performance was shown by a higher-ranking score.

Table 7 and Table 8 show that CDDO–HSA had the highest-ranking score of 2.5, followed by DCSO with a score of 2.8. FOX had a score of 3.1, which was higher than the scores of the other three algorithms: GWO–WOA (3.6), WOA–BAT (3.8), and ChOA (5.8). Finally, BOA had the lowest ranking score of six.

**Table 8.** The ranking score of CDDO–HS, ChOA, BOA, FOX, GWO–WOA, WOA–BAT, and DCSO algorithms.

| Function | CDDO–HS | ChOA | BOA | FOX | GWO–WOA | WOA–BAT | DCSO |
|---|---|---|---|---|---|---|---|
| Subtotal | 2.5 | 5.8 | 6 | 3.1 | 3.6 | 3.8 | 2.8 |

## 7. Conclusions

To summarize, both CDDO and global HS, together with their algorithmic features, were thoroughly discussed. The CDDO-HS technique was given. The experimental data were described to evaluate CDDO–HS's performance. Numerous tests were carried out to evaluate CDDO–HS. CDDO–HS was evaluated using 19 traditional benchmark test functions in both exploitation and exploration. When compared to CDDO and standard HS, CDDO–HS outperformed in 11 of 19 functions. CDDO–HS was also evaluated using CEC-C06 2019 benchmark functions. As a result, CDDO–HS performed admirably in seven of the ten functions. Despite conventional HS outperforming CDDO–HS in the remaining three functions, and having a better overall outcome, CDDO–HS outperformed standard HS in seven of the ten functions.

Additionally, CDDO–HS was statistically evaluated using the Wilcoxon rank-sum test, and it achieved overall significant results in the classical and CEC-C06 2019 benchmark functions. The CDDO–HS algorithm was then compared to the ChOA, BOA, FOX, GWO–WOA, WOA–BAT, and DCSO algorithms, and it outperformed all six metaheuristic approach algorithms in six of the ten benchmarks.

As a result, CDDO–HS's performance in terms of its exploration capability was enhanced. Generally, CDDO–HS improved the solution's quality with each iteration while avoiding the local optima.

Finally, the following potential research projects could be undertaken in the future:

1. Addressing real-world difficulties, such as medical, applied science, and engineering issues.
2. Mixing different methods to improve the results we already have.
3. Change the standard HS parameters to improve the exploration and exploitation phases and achieve better performance as a result.

**Author Contributions:** Conceptualization, A.A.A, T.A.R and S.A.; methodology, A.A.A, T.A.R and S.A.; software, A.A.A; validation, A.A.A. and T.A.R.; formal analysis, A.A.A. and T.A.R.; investigation, T.A.R; resources, A.A.A.; data curation, A.A.A.; writing—original draft preparation, A.A.A.; writing—review and editing, T.A.R.; visualization, A.A.A, T.A.R and S.A.; supervision, T.A.R and S.A.; project administration, T.A.R.; funding acquisition, A.A.A. All authors have read and agreed to the published version of the manuscript.

**Funding:** This research received no external funding.

**Conflicts of Interest:** The authors declare no conflict of interest.

## Appendix A

The mathematical formulation of the conventional benchmark functions utilized in this work is shown in Tables A1–A3.

**Table A1.** Unimodal test functions [72].

| Function | Dimension | Range | $f_{min}$ |
|---|---|---|---|
| $F_1(x) = \sum_{i=1}^{n} X_i^2$ | 10 | $[-100, 100]$ | 0 |
| $F_2(x) = \sum_{i=1}^{n} |X_i| + \prod_{i=1}^{n} |X_i|$ | 10 | $[-10, 10]$ | 0 |

| Function | | Dimension | Range | $f_{min}$ |
|---|---|---|---|---|
| $F_3(x) = \sum_{i=1}^{n}\left(\sum_{j=1}^{i} X_j\right)^2$ | | 10 | $[-30, 30]$ | 0 |
| $F_4(x) = max_i\{|X_i|, 1 \leq i \leq n\}$ | | 10 | $[-100, 100]$ | 0 |
| $F_5(x) = \sum_{i=1}^{n-1}[100(x_{i+1} - x_i^2)^2 + (x_i - 1)^2]$ | | 10 | $[-30, 30]$ | 0 |
| $F_6(x) = \sum_{i=1}^{n}([x_i + 0.5])^2$ | | 10 | $[-100, 100]$ | 0 |
| $F_7(x) = \sum_{i=1}^{n} ix_i^4 + random[0,1]$ | | 10 | $[-1.28, 1.28]$ | 0 |

**Table A2.** Multimodal test functions [72].

| Function | Dimension | Range | $f_{min}$ |
|---|---|---|---|
| $F_8(x) = \sum_{i=1}^{n} -X_i^2 \sin\left(\sqrt{|x_i|}\right)$ | 10 | $[-500, 500]$ | 0 |
| $F_9(x) = \sum_{i=1}^{n}[x_i^2 - 10\cos(2\pi x_i) + 10]$ | 10 | $[-10, 10]$ | 0 |
| $F_{10}(x) = -20exp\left(-0.2\sqrt{\sum_{i=1}^{n} x_i^2}\right) - exp\left(\frac{1}{n}\sum_{i=1}^{n}\cos(2\pi x_i)\right) + 20 + e$ | 10 | $[-32, 32]$ | 0 |
| $F_{11}(x) = \frac{1}{4000}\sum_{i=1}^{n} x_i^2 - \prod_{i=1}^{n}\cos\left(\frac{x_i}{\sqrt{i}}\right) + 1$ | 10 | $[-600, 600]$ | 0 |
| $F_{12}(x) = \frac{\pi}{n}\left\{10\sin(\pi y_1) + \sum_{i=1}^{n-1}(y_i - 1)^2[1 + 10\sin(\pi y_{i+1})^2] + (y_n - 1)^2\right\}$ $+ \sum_{i=1}^{n} \mu(x_i, 10, 100, 4), y_i = 1 + \frac{x+1}{4}, \mu(x_i, a, k, m) = \begin{cases} k(x_i - a)^m & x_i > a \\ 0 & -a < x_i < a \\ k(-x_i - a)^m & x_i < -a \end{cases}$ | 10 | $[-50, 50]$ | 0 |
| $F_{13}(x) = 0.1\left\{\sin(3\pi x_1)^2 + \sum_{i=1}^{n}(x_i - 1)^2[1 + \sin(3\pi x_i + 1)^2] + (x_n - 1)^2[1 + \sin(2\pi x_n)^2]\right\}$ $+ \sum_{i=1}^{n} \mu(x_i, 5, 100, 4)$ | 30 | $[150, 50]$ | 0 |

**Table A3.** Fixed-dimension multimodal benchmark functions.

| Function | Dimension | Range | $f_{min}$ |
|---|---|---|---|

| Function | Dim | Range | $f_{min}$ |
|---|---|---|---|
| $F_{14}(x) = \left(\dfrac{1}{500} + \sum_{j=1}^{25} \dfrac{1}{j + \sum_{i=1}^{2}(x_i - a_{ij})^6}\right)^{-1}$ | 2 | $[-65, 65]$ | 1 |
| $F_{15}(x) = \sum_{i=1}^{11}\left[a_i - \dfrac{x_1(b_i^2 + b_i x_2)}{b_i^2 + b_i x_3 + x_4}\right]^2$ | 4 | $[-5, 5]$ | 0.0003 |
| $F_{16}(x) = 4x_1^2 - 2.1x_1^4 + \dfrac{1}{3}x_1^6 + x_1 x_2 - 4x_2^2 + 4x_2^4$ | 2 | $[-5, 5]$ | $-1.0316$ |
| $F_{17}(x) = \left(x_2 - \dfrac{5.1}{4\pi^2}x_1^2 + \dfrac{5}{\pi}x_1 - 6\right)^2 + 10\left(1 - \dfrac{1}{8\pi}\right)\cos x_1 + 10$ | 2 | $[-5, 5]$ | 0.398 |
| $F_{18}(x) = [1 + (x_1 + x_2 + 1)^2(19 - 14x_1 + 3x_1^2 - 14x_2 + 6x_1 x_2 + 3x_2^2)] \times [30 + (2x_1 - 3x_2)^2 \times (18 - 32x_1 + 12x_1^2 + 48x_2 - 36x_1 x_2 + 27x_2^2)]$ | 2 | $[-2, 2]$ | 3 |
| $F_{19}(x) = -\sum_{i=1}^{4} c_i \exp\left(-\sum_{j=1}^{3} a_{ij}(x_j - p_{ij})^2\right)$ | 3 | $[1, 3]$ | $-3.86$ |